\documentclass{article}
\usepackage{spconf,amsmath,graphicx}
\usepackage{amssymb}
\usepackage{comment}
\usepackage{multirow}
\usepackage{multicol}
\usepackage{booktabs}
\usepackage{pifont}
\usepackage{makecell}
\newcommand{\cmark}{\ding{51}}

\usepackage{arydshln}

\makeatletter
\def\blfootnote{\xdef\@thefnmark{}\@footnotetext}
\makeatother

\title{Multi-Temporal Lip-Audio Memory for Visual Speech Recognition}
\name{Jeong Hun Yeo, Minsu Kim, Yong Man Ro$^\dagger$\thanks{This work was supported by IITP grant funded by the Korea government(MSIT) (No.2020-0-00004, Development of Previsional Intelligence based on Long-Term Visual Memory Network). $^\dagger$Corresponding author.}}
\address{
    School of Electrical Engineering, KAIST, South Korea\\
    \{sedne246, ms.k, ymro\}@kaist.ac.kr
}
\begin{document}
%
\maketitle
\begin{abstract}
Visual Speech Recognition (VSR) is a task to predict a sentence or word from lip movements. Some works have been recently presented which use audio signals to supplement visual information. However, existing methods utilize only limited information such as phoneme-level features and soft labels of Automatic Speech Recognition (ASR) networks. In this paper, we present a Multi-Temporal Lip-Audio Memory (MTLAM) that makes the best use of audio signals to complement insufficient information of lip movements. The proposed method is mainly composed of two parts: 1) MTLAM saves multi-temporal audio features produced from short- and long-term audio signals, and the MTLAM memorizes a visual-to-audio mapping to load stored multi-temporal audio features from visual features at the inference phase. 2) We design an audio temporal model to produce multi-temporal audio features capturing the context of neighboring words. In addition, to construct effective visual-to-audio mapping, the audio temporal models can generate audio features time-aligned with visual features. Through extensive experiments, we validate the effectiveness of the MTLAM achieving state-of-the-art performances on two public VSR datasets. 

\end{abstract}


\section{Introduction}
\label{sec:intro}
Visual Speech Recognition (VSR) \cite{ma2021end,prajwal2022sub, kim2022speaker} is a task to recognize words by interpreting the lip movements in a video without sound. It has recently received a lot of attention. VSR can be used as a subtitling tool for silent movies, as an ear for the deaf, and even as a tool to guess what the speaker is saying in noisy environments. However, inferring speech by watching lip movement only could be insufficient to model the speech content, due to human speech is not only produced by the lips but by the coordination of muscles and multiple organs \cite{sataloff1992human}.

Hence, researchers try to complement insufficient visual information from speech audio signals. A popular approach is exploiting audio knowledge via knowledge distillation \cite{zhao2020hearing,ren2021learningfromthemaster}, which consists of a visual student network and an audio teacher network. The visual student network is trained to follow the soft label or intermediate feature of the audio teacher network that pretrained with audio corpus data. However, due to the heterogeneity gap between visual and audio modalities, only limited information might be distilled by the audio network. To bypass the heterogeneity gap issue, VSR models augmented with memory networks have been proposed by \cite{kim2021cromm, kim2022mvm}. The memory network can bypass the heterogeneity gap by using constructed mapping between visual and audio, which can be utilized to load audio features without real audio input at the inference stage. However, the existing memory-augmented VSR models utilize only partial information of audio such as phoneme-level representations encoded from a short-term audio signal. Since short-term audio hardly contains context information, long-term audio signal that contains context information of neighboring words might be more suitable in constructing accurate visual-to-audio mapping.

\begin{figure*}[t!]
	\begin{minipage}[b]{1.0\linewidth}
		\centering
		\centerline{\includegraphics[width=18cm]{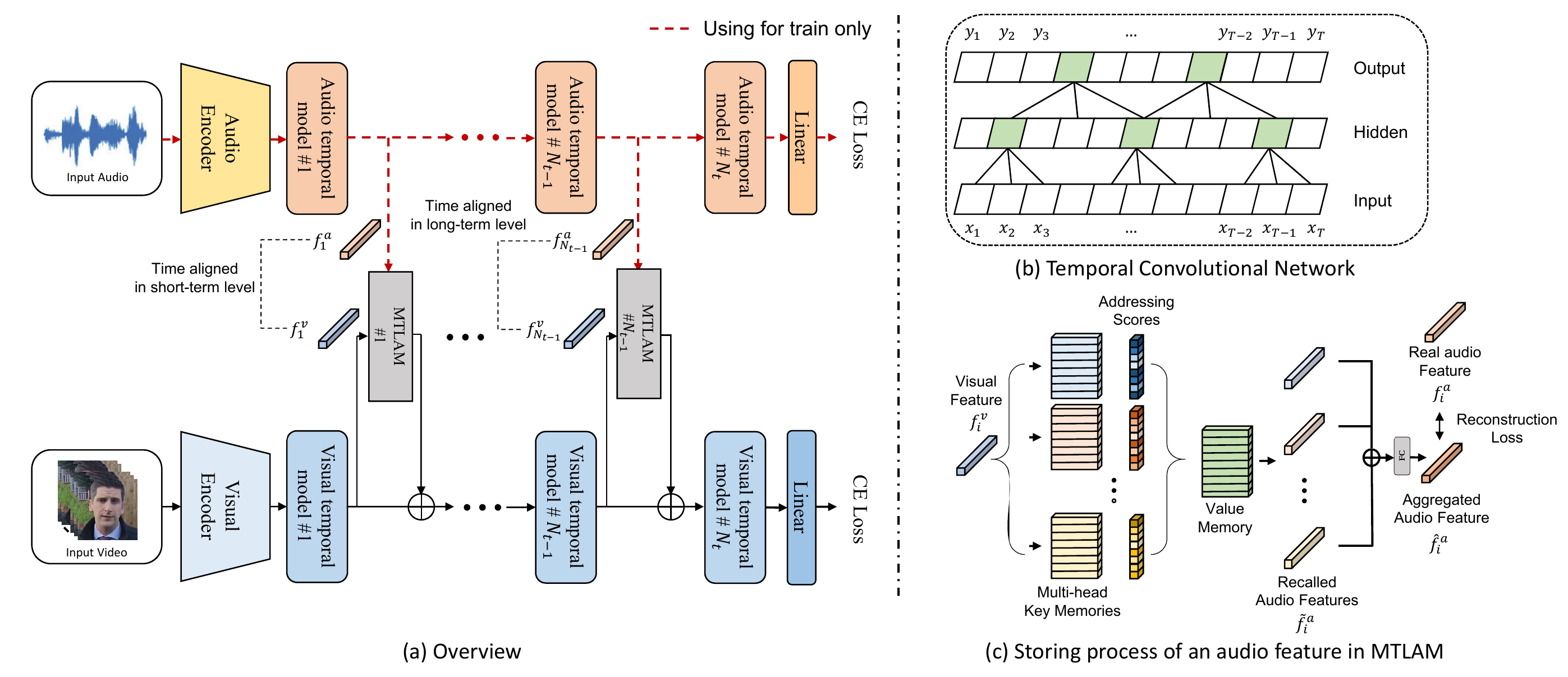}}
	\end{minipage}
	\caption{(a) Overall architecture of the proposed VSR framework with MTLAMs, (b) The temporal model gradually extends the temporal receptive field as the layer is deeper, (c) Storing process of MTLAM using multi-head key and one value memories.}
	\label{fig:1}
\end{figure*}

To address this problem, we present Multi-Temporal Lip-Audio Memory Networks (MTLAM) to complement insufficient visual information of the VSR model through multi-temporal audio features encoded from short- to long-term audio signals. To utilize audio features encoded in multi-temporal scales, we consider the following two key points: 1) An additional audio temporal model is required to capture context information from long-term audio signals. 2) The audio features encoded from each layer of the audio temporal model should be time aligned with each visual feature to construct effective visual-to-audio mapping in memory networks. To this end, we design an audio temporal model that can capture context from a sequence of audio features which has the same temporal information as the visual feature, so that we can obtain time-aligned audio features with visual features. The time-aligned audio features can improve accurate mapping of visual-audio features in memory. Finally, the MTLAMs store multi-temporal audio features encoded from audio temporal model so the multi-temporal context, from short to long-term, can be considered. We validate the effectiveness of MTLAM through two VSR benchmark datasets, namely, LRW and LRW-1000, and demonstrate the importance of utilizing context information of audio signals in constructing accurate visual-to-audio mapping. In addition, the proposed MTLAM achieves state-of-the-art performances on two public VSR benchmark databases.

\section{Method}
\label{sec:format}
\subsection{Overview}
The overall pipeline of the proposed method is depicted in Fig.~\ref{fig:1} (a). Firstly, to produce a spatio-temporal visual feature and audio feature from the lip movement video and raw audio signal, we employ a visual encoder and audio encoder, respectively. Next, we use the visual temporal model and audio temporal model to produce multi-temporal visual features and multi-temporal audio features having short- to long-term context information, and the generated multi-temporal visual features are complemented by multi-temporal audio features through MTLAMs. Finally, the complemented visual features are fed into a linear layer to predict ground-truth text corresponding to the input lip movements. In addition, to guide the audio features to have meaningful speech representations, we also train the audio stream to predict the text during training. 

\subsection{Audio Temporal Model}
Different from the previous work \cite{kim2022mvm} that utilized fixed short-term audio features, we build an additional audio temporal model to consider multi-temporal audio features having short- and long-term context information. The audio temporal model is composed of Temporal Convolution Network (TCN) \cite{lea2017temporal}, as shown in Fig. \ref{fig:1} (b). The TCN performs 1-D convolutions along to temporal dimension and captures context information from the sequence of features. The receptive field size $R$ of each TCN layer with stride 1 can be calculated as:
\begin{equation}
    R = (k-1)(d-1) + k,
\end{equation}
where $k$ is a filter size, and $d$ is a dilation size. For instance, the receptive field with filter size 3, dilation size 1, and stride 1 is 3, which is illustrated as hidden layer in Fig. \ref{fig:1} (b). By stacking TCN layers, we can extract a short- and long-term context from audio inputs through gradually increasing receptive fields. In addition, to construct effective visual-to-audio mapping in MTLAM, the audio temporal model is designed to produce audio features time-aligned with visual features. Specifically, we design TCN layers so that the receptive field of each layer of the audio temporal model is the same as each layer of the visual temporal model, thus the two modal representations are synchronized in time.

\subsection{Multi-Temporal Lip-Audio Memory Network}
We propose MTLAM to supplement the insufficient visual information through multi-temporal audio features that contain short- and long-term context of speech audio encoded with the audio temporal model. We denote the multi-temporal audio features encoded with $N_{t-1}$ layers of the audio temporal model except for the last layer as $\{f^a_i\}_{i=1}^{N_{t-1}}$, where $f^a_i \in \mathbb{R}^{T\times D}$, $T$ is the length of a sequence of audio features, and $D$ is the embedding dimension. Similarly, the multi-temporal visual features produced from visual temporal model denoted as $\{f^v_i\}_{i=1}^{N_{t-1}}$, where $f^v_i \in \mathbb{R}^{T\times D}$.

To store $N_{t-1}$ multi-temporal audio features, the MTLAM has architecture based on \cite{kim2022mvm} which consists of multi-head key memories and a value memory to consider visual-to-audio mapping as one-to-many mapping. Therefore, a total of $N_{t-1}$ multi-head key memories $M_{k, i} \in \mathbb{R}^{h \times N \times D}$ and $N_{t-1}$ value memories $M_{v, i} \in \mathbb{R}^{N \times D}$ are employed to construct visual-to-audio mapping at multi-temporal level in MTLAM, where the $h$ is the number of heads and the $N$ is the number of memory slots. 

Firstly, we construct $N_{t-1}$ multi-temporal visual-to-audio mapping through $N_{t-1}$ multi-head key memories in MTLAM. Specifically, each $j$-th slots of multi-head key memories are connected with each $j$-th slot of value memory. Therefore, the visual and audio representations from the same word or phrase can be mapped. To calculate all connections between key memories and the value memory, an addressing score $A_{k, i} \in R^{h \times N}$ is obtained with the following equation:
\begin{equation}
    A_{k,i} = \sigma (\alpha \cdot d(M_{k,i}, W_{q,i}^T f^v_{i})),
\end{equation}
where $d(\cdot)$ is cosine similarity, $\sigma$ indicate Softmax function, $\alpha$ is scaling factor, and $W_{q,i} \in R^{h \times D} $ is an embedding matrix. 

Then the saved audio features $\tilde{f}^a_{i} \in R^{h \times D}$ can be recalled by accessing the value memory using the addressing score $A_{k,i}$ as follows:
\begin{equation}
    \tilde{f}^a_{i} = A_{k,i} \cdot M_{v,i}.
\end{equation}
The $h$ recalled audio features are aggregated into a single audio feature $\hat{f}^a_{i} \in \mathbb{R}^{T\times D}$ time aligned with the input visual feature through an aggregation layer $W_a \in \mathbb{R}^{Dh \times D}$. This procedure is shown in Fig. \ref{fig:1} (c). Then, the aggregated multi-temporal audio features $\hat{f}^a_i$ are fused with multi-temporal visual features $f^v_i$ in each layer through summation to complement the visual speech information.

\subsection{Joint Loss Function}
The proposed method utilizing multi-temporal audio features is jointly trained in an end-to-end manner. Firstly, to store and reconstruct multi-temporal audio features, we employ a reconstruction loss and contrastive loss. The reconstruction loss is designed to reduce distances between reconstructed multi-temporal audio features and original multi-temporal audio features, which can be formulated as: 
\begin{equation}
    \mathcal{L}_{recon} = \frac{1}{N_{t-1}} \sum_{i=1}^{N_{t-1}} ||1 - d(\hat{f}^a_{i},f^a_{i})||_1. 
\end{equation}
Next, the contrastive loss is employed to store different audio features into different slots of value memories, which can be formulated as:
\begin{equation}
    \mathcal{L}_{cont} = \frac{1}{N_{t-1}}\sum_{i=1}^{N_{t-1}} (\sum_{p\neq q} d(M_{v,i}^p,M_{v,i}^q) + \sum_{p=q} d(M_{v,i}^p,M_{v,i}^q)).
\end{equation}
Finally, to train visual and audio-models, two classification losses are used for VSR and ASR. $\mathcal{L}_{cls}^v = \mathcal{L}_{CE}(\hat{y}_v, y)$, $\mathcal{L}_{cls}^a = \mathcal{L}_{CE}(\hat{y}_a, y)$, where the $\hat{y}_v$ and $\hat{y}_a$ are predicted labels from the VSR network and ASR network, respectively. The $y$ is a ground truth. In addition, a classification loss $\mathcal{L}_{cls}^{v,\hat{a}} = \mathcal{L}_{CE}(\hat{y}_{v,\hat{a}}, y)$ is designed to train the proposed MTLAM, where $\hat{y}_{v,\hat{a}}$ is predicted label using visual feature with reconstructed multi-temporal audio features. Therefore, the total loss can be formulated as:
\begin{equation}
    \mathcal{L}_{total} =  \mathcal{L}_{cls}^v + \mathcal{L}_{cls}^a + \mathcal{L}_{cls}^{v,\hat{a}} +\mathcal{L}_{recon} + \mathcal{L}_{cont}.
\end{equation}

\begin{table}[t]
\renewcommand{\arraystretch}{1.2}
\renewcommand{\tabcolsep}{3.0mm}
\caption{Effectiveness of each audio feature generated from each layer of audio temporal model}
\centering
\resizebox{0.8\linewidth}{!}{
\begin{tabular}{ccccc}
\Xhline{3\arrayrulewidth}
\multirow{2}{*}{\textbf{Baseline\cite{ma2022training}}} & \multicolumn{3}{c}{\textbf{MTLAM}} & \multirow{2}{*}{\textbf{Top-1 ACC. (\%)}}  \\
& \#1 & \#2  & \#3  \\ \hline
\cmark& - & - & - & 89.6\\ 
\cmark&\cmark & - & - & 90.9 \\
\cmark& &\cmark & - & 91.1\\
\cmark& - & - &\cmark & 90.8\\
\cmark&\cmark &\cmark & - & 91.3 \\
\cmark& - &\cmark &\cmark &  91.4\\ \hdashline
\cmark&\cmark &\cmark &\cmark & \textbf{91.7} \\
\Xhline{3\arrayrulewidth}
\end{tabular}}
\label{tab:1}
\end{table}


\section{Experimental Results}
\label{sec:pagestyle}
\subsection{Experimental setup}
\noindent\textbf{Dataset.} LRW \cite{chung2016lrw} is an English word-level VSR dataset which contains 500 words with 800$\sim$1,000 video clips per class. Each video clip has 29 frames with a 1.16 second duration. We preprocess the video to focus on lip movement. The video data is cropped into 136${\times}$136 pixels centered at the lip without aligning the face and resized into 96${\times}$96 pixels. The sampling rate of audio signals is 16,000Hz

LRW-1000 \cite{yang2019lrw1000} is a Mandarin word-level VSR dataset containing 1,000 words and a total of 718,808 video clips. The video of LRW-1000 is already cropped, and we only resize the video clip into 96${\times}$96 pixels. To match the time alignment between video and audio, we adjust the duration of the video clip to be 0.2 seconds longer.

\noindent \textbf{Implementation Details.} The visual encoder is consisted of ResNet-18 and 3D convolutional network, and the visual temporal model is composed of 4 DC-TCN \cite{ma2021denselyconnected} layers. Similarly, we use combination of ResNet-18 and 1D covolutional network as an audio encoder, and the audio temporal model has same structure with visual temporal model. To store multi-temporal audio features, we use 3 MTLAM between the visual- and audio-temporal model except for the last DC-TCN layer. We use eight NVIDIA GeForce RTX 3090 GPUs for training models. It takes about seven days to train a single model in an end-to-end manner for 80 epochs. The batch size of 32 and a cosine annealing strategy are used.

\begin{table}[t]
\renewcommand{\arraystretch}{1.2}
\renewcommand{\tabcolsep}{4.5mm}
\caption{Comparison with existing VSR methods}
\centering
\resizebox{0.99\linewidth}{!}{
\begin{tabular}{ccc}
\Xhline{3\arrayrulewidth}
\multirow{2}{*}{\textbf{Method}} & \multicolumn{2}{c}{\textbf{Top-1 ACC. (\%)}} \\ \cline{2-3}
& LRW  & LRW-1000  \\ \hline
Multi-Grained + C-BiLSTM \cite{wang2019multigrained} & 83.3 & 36.9 \\
R34 + BGRU \cite{yang2019lrw1000} & - & 38.2 \\
R18 + BGRU + GRU \cite{luo2020pcpg} & 83.5 & 38.7 \\
T I3D + BiLSTM \cite{weng2019twostream} & 84.1 & - \\
T R18 + BiGRU \cite{xiao2020deformation} & 84.1 & 41.2 \\
R18 + BiGRU + LSTM  \cite{zhao2020mi} & 84.4 & 38.8 \\
P3D R50 + BiLSTM \cite{xu2020discriminative} & 84.8 & - \\
R18 + BiGRU / Face Cutout \cite{zhang2020facecutout} & 85.0 & 45.2 \\
R18 + MS-TCN \cite{martinez2020mstcn} & 85.3 & 41.4 \\
R18 + MS-TCN / Born-Again \cite{ma2021towards} & 87.9 & 46.6 \\ 
R18 + MS-TCN / LiRA \cite{ma2021lira} & 88.1 & - \\
R18 + MS-TCN / WPCL + APFF \cite{tian2022lipreading} & 88.3 & - \\
R18 + DC-TCN \cite{ma2021denselyconnected} & 88.4 & 43.7 \\ 
R18 + MS-TCN / MVM \cite{kim2022mvm} & 88.5 & 53.8 \\
R18 + MS-TCN / NetVLAD \cite{yang2022improved} & 89.4 & - \\ 
 
R18 + DC-TCN / Training strategy \cite{ma2022training} & 91.6$^*$ & - \\ \hline
Proposed method(MTLAM) & \textbf{91.7} & \textbf{54.3} \\ 
\Xhline{3\arrayrulewidth}
\end{tabular}}
\label{tab:2}
\end{table}

\subsection{Results}
\noindent\textbf{Ablation Study.} The proposed MTLAM is designed to utilize multi-temporal audio features having context information of short- to long-term audio signals. The deeper the MTLAM, the longer the audio knowledge is stored. To validate the effectiveness of multi-temporal audio features stored in MTLAMs, we conduct experiments with 6 variant models on LRW dataset, and the results are shown in Table \ref{tab:1}. The baseline VSR model \cite{ma2022training} achieves 89.6\% word accuracy. When we utilize only one MTLAM in different temporal levels, MTLAM\#1, MTLAM\#2, and MTLAM\#3 achieve 90.9\%, 91.1\%, and 90.8\% word accuracies, respectively. The results show that audio knowledge saved in MTLAM\#2 is the most effective when the VSR model saves audio features of one temporal level into MTLAM. When using two MTLAMs, MTLAM\#1 with MTLAM\#2 achieves 91.3\%, and MTLAM\#2 with MTLAM\#3 achieves 91.4\% accuracy. These results show that using multi-temporal audio features is better than using a single MTLAM. Finally, using all of the multi-temporal audio features gives the best accuracy of 91.7\% by considering short- to long-term context, which is the final proposed method.

\blfootnote{$^*$ Ensemble performance of 4 trained models.}

\noindent\textbf{Comparison with state-of-the-art methods.}
To validate the effectiveness of our proposed method, we compare performances with the existing state-of-the-art models on LRW and LRW-1000 datasets. Table \ref{tab:2} shows the comparison results. Our proposed method achieves accuracy of 91.7\% and 90.1\% on LRW by using temporal models of DC-TCN and MS-TCN, respectively, and achieves the best performance among the baselines. Especially, the proposed method shows even better performance than one of the state-of-the-art method \cite{ma2022training} that used ensemble of 4 different trained models. In addition, we achieve 54.3\% word accuracy on LRW-1000. It shows that utilizing multi-temporal audio features having both short- and long-term context information for constructing visual-to-audio mapping in memory networks can improve the VSR performances.

\begin{figure}[t!]
	\begin{minipage}[b]{1.0\linewidth}
		\centering
		\centerline{\includegraphics[width=9cm]{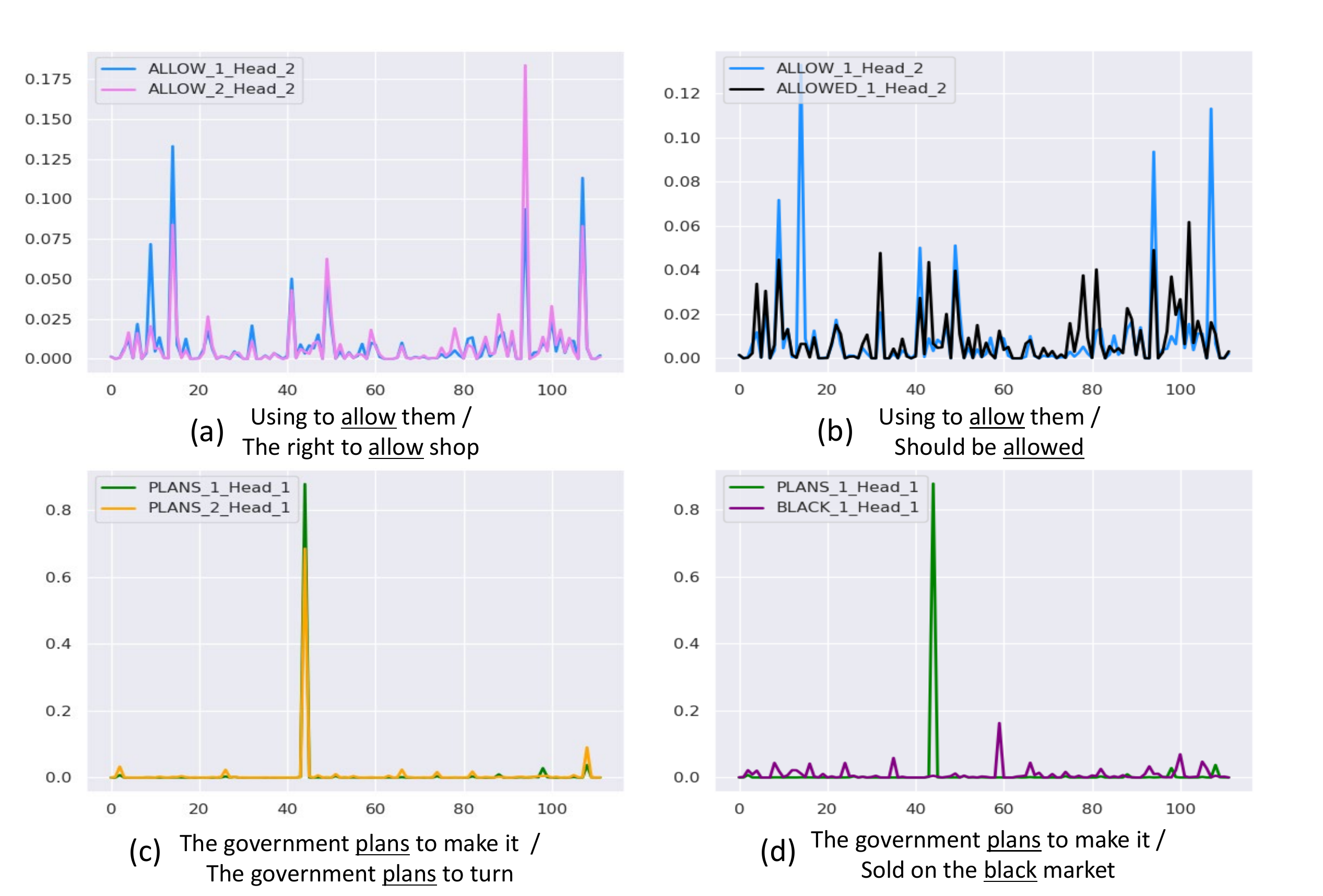}}
	\end{minipage}
	\caption{(a) and (c) are addressing scores when the audio pairs have similar context information, and (b) and (d) are addressing scores when the audio pairs have different context information.}
	\label{fig:2}
\end{figure}

\noindent\textbf{Long-term audio knowledge in MTLAM.}
To verify whether audio features with long-term context information are well stored in MTLAM, we visualize the addressing score of MTLAM\#2 of four different audio pairs. Fig. \ref{fig:2} (a) and (c) show the addressing scores when the audio pairs contain the same phrase and context in sentences. In contrast, the addressing scores of audio pairs having different phrase and context information are shown in Fig. \ref{fig:2} (b) and (d). The more similar the two sentences, the more consistent the distribution of addressing scores. This shows that the proposed MTLAM can also construct visual-to-audio mapping in a long-term context, in addition to a short-term context.

\section{Conclusion}
In this paper, we presented Multi-Temporal Lip-Audio Memory (MTLAM) utilizing multi-temporal audio knowledge to complement insufficient visual information in VSR. In addition, the audio temporal model for MTLAM generated multi-temporal audio features time-aligned with visual features to effectively map between visual- and audio features. We showed that the mapping between visual and audio features can be enhanced by considering the multi-temporal context in audio and visual features. We validated the effectiveness of the proposed method by achieving state-of-the-art performances on two VSR benchmark databases.

\clearpage
\vfill\pagebreak

{\small
\bibliographystyle{IEEEbib}
\bibliography{main}
}
\end{document}